\documentclass[fleqn,10pt]{wlscirep}
\usepackage[utf8]{inputenc}
\usepackage[T1]{fontenc}
\usepackage{lineno}
\usepackage{graphicx}
\graphicspath{ {./figures/} }

\title{The RW3D: A multi-modal panel dataset to understand the psychological impact of the pandemic}

\author[1,3$,\dag$]{Isabelle van der Vegt}
\author[2,3,$\dag$]{Bennett Kleinberg}
\affil[1]{Utrecht University, Department of Sociology, Utrecht University, 3584 CH Utrecht, The Netherlands}
\affil[2]{Tilburg University, Department of Methodology and Statistics, 5037 AB Tilburg, The Netherlands}
\affil[3]{University College London, Department of Security and Crime Science, London WC1E 6BT, United Kingdom}
\affil[*]{corresponding author: Isabelle van der Vegt (i.w.j.vandervegt@uu.nl)}

\affil[$\dag$]{these authors contributed equally to this work}

\begin{abstract}
    Besides far-reaching public health consequences, the COVID-19 pandemic had a significant psychological impact on people around the world. To gain further insight into this matter, we introduce the Real World Worry Waves Dataset (RW3D). The dataset combines rich open-ended free-text responses with survey data on emotions, significant life events, and psychological stressors in a repeated-measures design in the UK over three years (2020: $n=2441$, 2021: $n=1716$ and 2022: $n=1152$). This paper provides background information on the data collection procedure, the recorded variables, participants' demographics, and higher-order psychological and text-based derived variables that emerged from the data. The RW3D is a unique primary data resource that could inspire new research questions on the psychological impact of the pandemic, especially those that connect modalities (here: text data, psychological survey variables and demographics) over time. 
\end{abstract}

\begin{document}

\flushbottom
\maketitle

\thispagestyle{empty}

\section*{Background \& Summary}
Since the start of the pandemic, social and behavioural scientists have collected data on the psychological impact on individuals of COVID-19 and the measures introduced around it. The global health crisis severely impacted lives around the world. At the same time, it enabled social scientists across various domains to study the response of humans to unprecedented circumstances. Several papers and associated datasets have emerged as a result of this, including those that adopted a psychological perspective. For instance, the \textit{COVIDiSTRESS Global Survey} includes measures such as perceived stress, trust in authorities, and compliance with anti-covid measures collected between 30 March and 30 May 2020 from 173,426 individuals across 39 countries and regions \cite{yamada_covidistress_2021}. Similarly, the \textit{PsyCorona} dataset consists of data collected at the start of the pandemic ($n = 34,526$) from 41 societies worldwide, measuring psychological variables and behaviours such as leaving the home and physical distancing \cite{kreienkamp_psycorona_2020}. That dataset has been used in follow-up studies to measure, for example, cooperation and trust across societies \cite{romano_cooperation_2021} and associations between emotion and risk perception of COVID-19 \cite{han_associations_2021}. Others have studied the concept of ‘pandemic fatigue’ (i.e., the perceived inability to “keep up” with restrictions), for which there are data available from eight countries \cite{jorgensen2022pandemic}. Associations between pandemic fatigue and the severity of restrictions were found, in addition to pandemic fatigue eliciting political discontent. 

Of particular promise to understand how individuals fared during and in the aftermath of the pandemic are free-text responses, which allow for more depth and coverage of topics than targeted survey-style data collection. Some initiatives have used and made available linguistic data on the consequences of the pandemic, usually in the form of Twitter data \cite{banda_large-scale_2021,naseem_covidsenti_2021}. In another study, Reddit and survey data were analysed to measure shifts in psychological states throughout the pandemic \cite{ashokkumar_social_2021}. However, both forms of data were collected from different participants, which does not allow for deeper exploration of ground truth psychological states of text authors by connecting survey and text modalities. Collecting text and survey data from the same participants is desirable for several reasons. Firstly, free-text responses enable participants to report their experiences in the pandemic in an unconstrained manner, potentially offering deeper insight into psychological processes. Second, simultaneously obtained survey responses offer ground truth measures on the psychological variables potentially underlying what is written about in text. Third, advances made in the area of natural language processing allow for in-depth quantitative analyses of the text data, thereby making text data a resource that reaches beyond qualitative analyses typically conducted manually. However, collecting data that connects the textual dimension to survey data is costly as it requires primary data collection and cannot be realised through “found data” (e.g., posts on social media). Consequently, to date, such datasets are scarce and the lack thereof has impeded how we can study the psychological impact of the pandemic.

The current paper fills that gap and introduces the \textit{Real World Worries Wave Dataset (RW3D)}, which offers the unique combination of ground-truth survey data on emotions with free-text responses describing emotions in relation to the pandemic. The richness of this dataset allows us to examine, for example, emotional responses, coping styles, and the content of worries as a consequence of COVID-19. Due to the broad scope of potential research questions and the scarcity and necessity of these data sources, we make this dataset available to the research community. Hereafter, we provide detailed background on the data collection procedure, recorded variables, participant demographics as well as an attrition analysis and descriptive statistics. Our aim with this paper is to offer detail on a unique resource that could inspire plenty of research questions.

\section*{Methods}

\subsection*{Ethics}
The data collection was approved by the departmental IRB at University College London. No personal data were collected from participants and all participants provided informed consent for participation and for their data to be shared for research purposes.

\subsection*{Procedure}
The dataset was collected in three waves\footnote{Note that we use wave in the study design sense, not to be confused with waves in the COVID-19 pandemic} in April of 2020, 2021 and 2022. Data collection started in April 2020 on the crowdsourcing platform Prolific Academic with an initial sample size of $n=2500$. We then contacted the same participants through the crowdsourcing platform one year later about a follow-up data collection and made participation slots available for all participants whose data were collected in the first wave. That procedure was repeated another year later with those participants whose data were collected in wave 2.

\begin{figure}
\centering
\includegraphics[width=\linewidth]{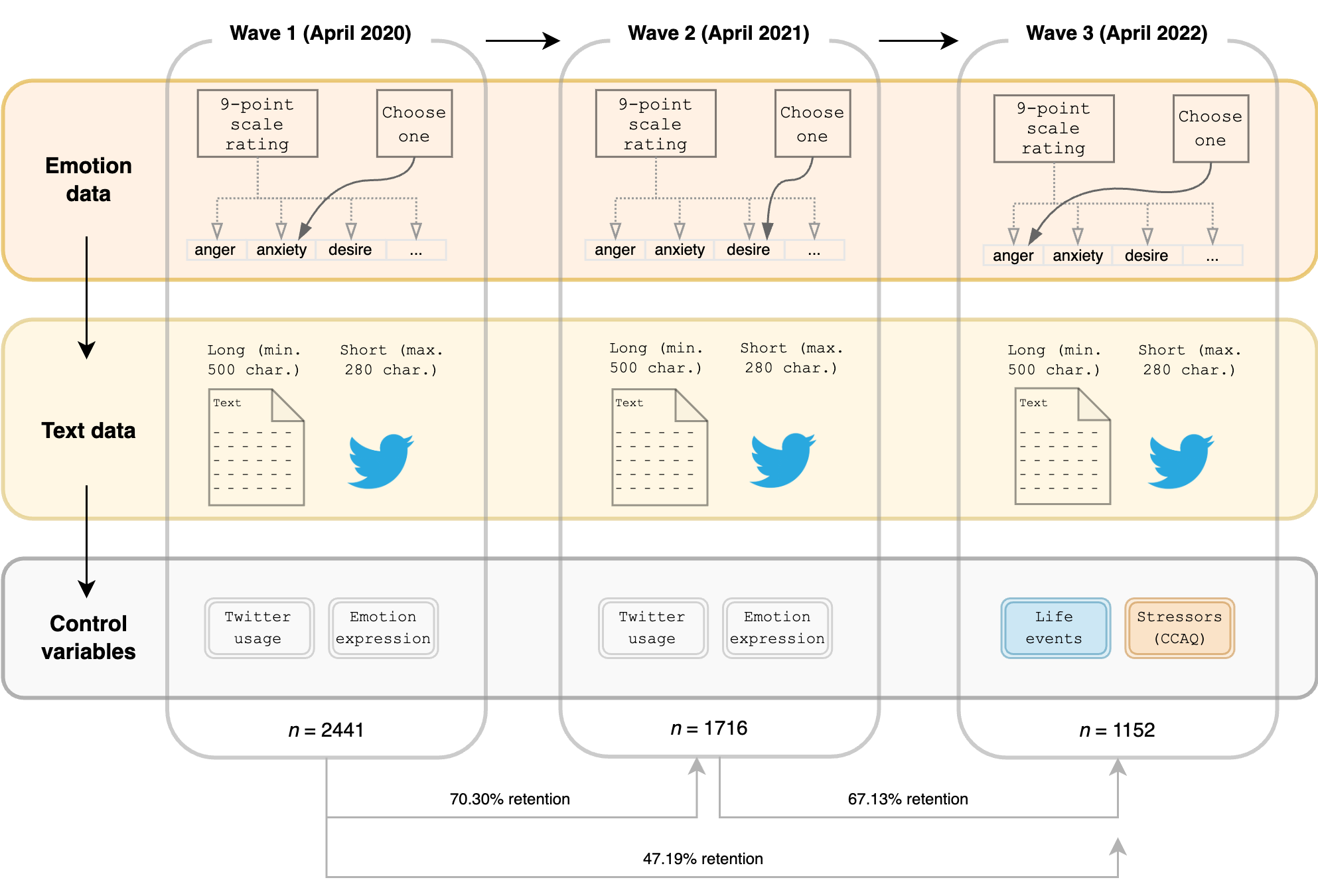}
\caption{Study design of three waves of data collection. Participants first provided emotion data in the form of 9-point scale ratings of eight emotions and selected the single best-fitting one, then wrote a long and a short text, before they filled in questionnaires about control variables (wave 1 and 2) or provided details on their life events and psychological stressors (wave 3). Demographic variables were obtained from Prolific.}
\label{fig1}
\end{figure}

In all data collection phases, participants were informed about the nature of the study. Participants started with the self-rated emotions questionnaire and the single emotion selection, then proceeded to the textual expressions and finally provided control variables (wave 1 and 2) and life events and psychological stressor variables (wave 3 only, see Fig. \ref{fig1}).

Only UK-based Prolific users who used Twitter as per the website’s prescreener\footnote{Here: those who selected Twitter for “Which of the following social media sites do you use on a regular basis (at least once a month)?”} were eligible for participation. Upon completion of the task, each participant was paid GBP 0.50. Even though the effective time spent on the task was somewhat longer than anticipated, we did not adjust the rewards so as not to introduce a change in reward as a confounding variable for the repeated-measures design. The task was administered through Qualtrics.

\subsection*{Timing and societal context}
The first wave of data collection occurred in early April 2020, when the UK was under lockdown with death tolls increasing. Queen Elizabeth II had just addressed the nation and Prime Minister Boris Johnson was admitted to hospital due to COVID-19 symptoms (see \cite{kleinberg2020measuring}). In wave 2 (April 2021), many people in the UK had been vaccinated, and schools, retail and the hospitality sectors were (partially) re-opening. The delta-variant of the Coronavirus had just been identified at this time (see \cite{mozes2021repeated}). Finally, in wave 3 (April 2022) all travel restrictions for those entering the UK had been lifted, the Omicron variant was surging and news around the Partygate affair (i.e., a political scandal surrounding parties held at Downing Street during lockdown) was ongoing \cite{strick_partygate_2022}. 

\begin{table}[!ht]
    \centering
    \begin{tabular}{|l|l|l|l|l|l|}
    \hline
        \textbf{Wave} & \textbf{Initial} & \textbf{Age} & \textbf{Female (\%)} & \textbf{Final} & \textbf{Retention (\%)} \\ \hline
        2020 & 2500 & 33.84 & 65.15 & 2441 & 100.00 \\ \hline
        2021 & 1839 & 36.22 & 67.37 & 1716 & 70.30 \\ \hline
        2022 & 1227 & 37.10 & 68.40 & 1152 & 67.13 \\ \hline
    \end{tabular}
    \caption{\label{tab:example}Basic sample characteristics per wave.}
    \label{t1}
\end{table}

\subsection*{Demographic variables}

We obtained participants’ demographics from Prolific. These are data that registered participants volunteered to provide and consist of their age, gender, country of birth, nationality, first language, employment status, student status, country of birth, country of residence as well as their participation on the crowdsourcing platform (number of tasks completed and approved). We have added one demographic question in the survey about their native language (as this may differ from their first language).

Participants were on average 37.10 years old ($SD=11.98$) in April 2022, of which 68.40\% were female (31.42\% male, remaining: prefer not to say), see Table \ref{t1}. The vast majority (90.54\%) indicated the UK as their country of birth and as their current country of residence (99.65\%). The latter observations match the recruitment pre-selection that we made. Regarding their employment status, in 2020, 52.43\% indicated being full-time employed, 22.74\% in part-time work and 10.50\% not in paid work (e.g., retired). Interestingly, the percentage of people in full-time work decreased somewhat in 2022 (42.35\%). Similarly, the percentage of students decreased from 16.93\% in 2020 to 10.94\% in 2022.

\subsection*{Emotion data}

\subsubsection*{Self-rated emotions} 
Participants were asked to indicate on a 9-point scale how worried they were about the Corona situation (1=not worried at all; 5=moderately worried; 9=very worried) and how they felt at this moment about the Corona situation. For the latter, they indicated how strongly they felt each of the following eight emotions (1=none at all; 5=moderately; 9=very much): anger, disgust, fear, anxiety, sadness, happiness, relaxation, desire \cite{harmon2016discrete}. The scale judgments were indicated using a slider in steps of 1.

\subsubsection*{Single emotion selection}
Of the eight emotions listed above (i.e., excluding worry), each participant was asked "If you have to choose just one, which of the emotions below best characterises how you feel at this moment?".

Table \ref{t2} shows the descriptive statistics for the emotion variables (self-rated scale values and discrete choice). While the pattern overall suggests improvement in that the positive emotions increase while the negative ones decrease, there are hidden patterns at play. Previous work found clusters of participants in how they transitioned from 2020 to 2021 \cite{mozes2021repeated} and we provide additional evidence for sub-groups below. 

\begin{table}[!ht]
    \centering
    \begin{tabular}{|l|l|l|l|l|l|l|l|l|l|}
    \hline
        \textbf{Emotion} & \textbf{$M_1$} & \textbf{$SD_1$} & \textbf{$prop._1$} & \textbf{$M_2$} & \textbf{$SD_1$} & \textbf{$prop._2$} & \textbf{$M_3$} & \textbf{$SD_3$} & \textbf{$prop._3$} \\ \hline
        worry & 6.67 & 1.70 & - & 5.07 & 2.03 & - & 3.98 & 2.16 & - \\ \hline
        anger & 3.76 & 2.18 & 0.04 & 3.47 & 2.35 & 0.08 & 2.85 & 2.19 & 0.06 \\ \hline
        disgust & 3.06 & 2.12 & 0.01 & 2.79 & 2.16 & 0.02 & 2.39 & 2.03 & 0.03 \\ \hline
        fear & 5.63 & 2.30 & 0.09 & 3.77 & 2.30 & 0.02 & 2.85 & 2.05 & 0.02 \\ \hline
        anxiety & 6.51 & 2.30 & 0.58 & 5.05 & 2.52 & 0.36 & 4.09 & 2.45 & 0.30 \\ \hline
        sadness & 5.55 & 2.31 & 0.15 & 4.64 & 2.57 & 0.19 & 3.48 & 2.35 & 0.13 \\ \hline
        happiness & 3.55 & 1.84 & 0.01 & 4.29 & 1.98 & 0.05 & 4.76 & 2.10 & 0.07 \\ \hline
        relaxation & 3.83 & 2.05 & 0.12 & 4.54 & 2.25 & 0.23 & 5.14 & 2.35 & 0.38 \\ \hline
        desire & 2.73 & 1.90 & 0.01 & 3.42 & 2.19 & 0.05 & 3.22 & 2.09 & 0.02 \\ \hline
    \end{tabular}
       \caption{\label{tab:example}Descriptive statistics per wave (M, SD) for the self-rated emotions (scale: 1=not at all; 5=moderately; 9=very much) and the proportion of individuals who chose the respective emotion as "best fitting" emotion.}
       \label{t2}
\end{table}

\subsection*{Text data}
We elicited two textual responses from each participant. The first text data was obtained through the following instruction: "Please write in a few sentences how you feel about the Corona situation at this very moment. This text should express your feelings at this moment." Participants typed their response in a text field and received a prompt if their response was shorter than 500 characters. The second text response was obtained directly thereafter aimed at eliciting a shorter, Tweet-length text as follows: "Suppose you had to express your current feeling about the Corona situation in a Tweet (max. 280 characters). Please write in the text box below". In this case, the participants were prompted if their text input was shorter than 10 or longer than 280 characters.

\begin{table}[!ht]
    \centering
    \begin{tabular}{|l|l|l|l|l|l|l|l|l|l|}
    \hline
        \textbf{Variable} & \textbf{$M_1$} & \textbf{$SD_1$} & \textbf{$range_1$} & \textbf{$M_2$} & \textbf{$SD_2$} & \textbf{$range_2$} & \textbf{$M_3$} & \textbf{$SD_3$} & \textbf{$range_3$} \\ \hline
        Long: Tokens & 126.17 & 39.37 & [62, 1067] & 125.17 & 32.05 & [78, 387] & 122.58 & 27.64 & [62, 358] \\ \hline
        Long: Chars. & 624.20 & 196.66 & [500, 5454] & 621.27 & 158.19 & [500, 2023] & 610.82 & 133.98 & [500, 1796] \\ \hline
        Short: Tokens & 26.63 & 15.90 & [1, 75] & 24.70 & 15.40 & [1, 75] & 24.59 & 14.62 & [1, 67] \\ \hline
        Short: Chars. & 131.11 & 76.97 & [10, 281] & 121.64 & 75.12 & [10, 281] & 122.01 & 73.36 & [10, 282] \\ \hline
    \end{tabular}
      \caption{\label{tab:example}Corpus descriptives (M, SD, range) for all three waves of data collection. Note that the exceeding of 280 characters in the short texts is due to differences in counting leading and trailing white spaces}
      \label{t3}
\end{table}

\begin{figure}[!ht]
\centering
\includegraphics[width=\linewidth]{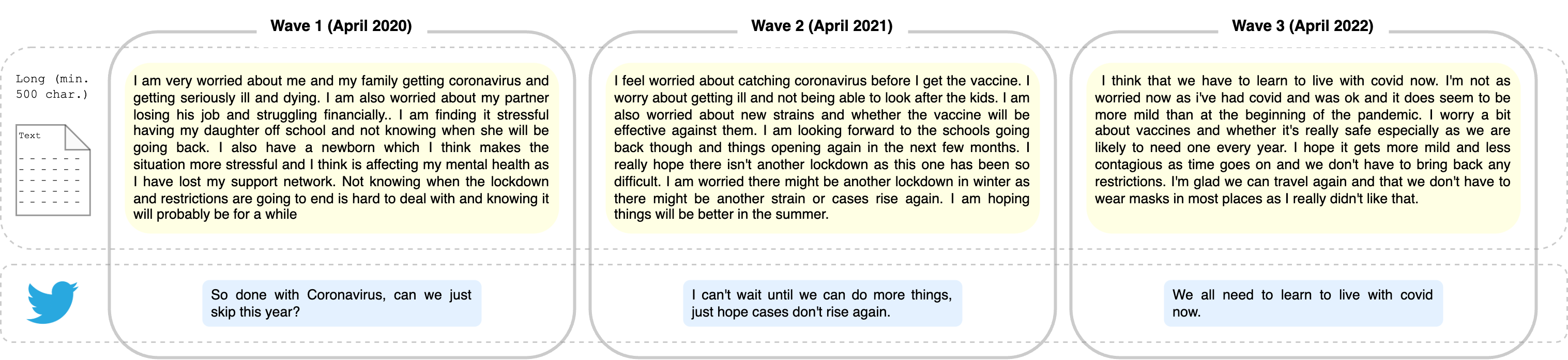}
\caption{Text data of a single participant (long text and Tweet-size text)}
\label{fig2}
\end{figure}

The corpus descriptives (Table \ref{t3}) show a stable length of both long and short texts over the three waves. In total, the corpus consists of 430,751 tokens (2020: 145,348; 2021: 144,191; 2022: 141,212). Below, text examples show the texts written by two participants over all three waves (Fig. \ref{fig2}).

\subsection*{Control variables in wave 1 and 2}
In the first two waves (April 2020 and 2021), we recorded two sets of control variables: the self-rated ability to express emotions in text and Twitter usage. We decided to drop these from the third wave. The rationale for dropping these variables was that we assumed these to change little within the individual and we already had two measurements (wave 1 and 2) that correlated substantially (supplementary materials Table 1).

\subsubsection*{Emotion expression}
As a potential control for the link between self-reported emotions on a survey and the expression of emotion in text, we asked participants to indicate on a 9-point scale (1=not at all; 5=moderately; 9=very well) how well they i) could express their feelings in general, ii) how well in the Tweet-size text, and iii) how well in the longer text.

\subsubsection*{Twitter usage}
As an additional potential confounding variable specifically for the Tweet-size text we asked about participants' Twitter usage. Using a 9-point scale (1=never; 5=every month; 9=every day), participants indicated how often they i) are on Twitter, ii) send Tweets themselves, and iii) participate in conversations on Twitter.

\subsection*{New variables in wave 3}
The most recent wave (April 2022) included two additional constructs that replaced the control variables from the previous waves. To better understand potential moderating variables of participants' emotional adjustment in the pandemic and their textual expression, we collected data on important life events during the pandemic and used a crisis coping questionnaire \cite{ccaq}.

\subsubsection*{Life events}
Participants were asked retrospectively about any important events or changes in their life that have happened to them over the past two years. First, they were asked whether "anything - positive or negative - in [their] life [has] over the past two years impacted how [they] dealt with the Corona situation?" Those who answered yes were then asked to describe the event, date the event (month and year) and rate the event's impact on a scale from -10 (very negative) to +10 (very positive). If there was an additional event, participants could also submit one more (for a maximum of two events). All life events were subsequently qualitatively coded by the authors to arrive at overarching categories. For instance, being fired, changing jobs, and obtaining a first job after college were mapped to the category ‘job’; getting married, finding a partner, and a break-up were all mapped to ‘romantic’ (see supplemental materials Table 4 for further examples). 

A third of the participants (33.85\%) reported a significant life event during data collection. The most common life event category was ‘death’ (e.g., a death in the family), which was almost exclusively rated as a negative life event (97.59\%). Life events related to work (e.g., a job change) were also common, which most participants (69.88\%) rated with a positive intensity (Table \ref{t5}). Other life events such as ‘mental health’ (e.g., experiencing panic attacks, receiving a mental health diagnosis) and ‘financial’ (e.g., paying off loans, loss of income) show a more ambivalent pattern and were rated as positive and negative with approximately equal proportion. Most life events occurred in December 2021 (median). See supplemental materials Table 4 for examples of each life event category. 

\begin{table}[!ht]
    \centering
    \begin{tabular}{|l|l|l|l|l|l|}
    \hline
        \textbf{Event} & \textbf{Prop.} & \textbf{M (SD) intensity} & \textbf{Median intensity} & \textbf{Prop. neg. intensity} & \textbf{Prop. pos. intensity} \\ \hline
        no life event & 66.15 & NA & NA & NA & NA \\ \hline
        death & 7.20 & -8.43 (2.83) & -10 [-10;6] & 97.59 & 2.41 \\ \hline
        job & 7.20 & 3.08 (6.3) & 6 [-10;10] & 28.92 & 69.88 \\ \hline
        romantic & 3.12 & -0.33 (7.95) & -4 [-10;10] & 52.78 & 47.22 \\ \hline
        family & 3.04 & -1.88 (7.48) & -6 [-10;10] & 65.71 & 37.14 \\ \hline
        reproduction & 3.04 & 5.94 (5.97) & 8 [-10;10] & 17.14 & 82.86 \\ \hline
        health & 2.52 & -5 (6.38) & -8 [-10;10] & 75.86 & 24.14 \\ \hline
        move & 2.34 & 5.85 (4.93) & 8 [-8;10] & 11.11 & 88.89 \\ \hline
        health of family & 1.91 & -7.91 (2.58) & -8 [-10;0] & 95.45 & 0.00 \\ \hline
        mental health & 1.48 & -1.18 (7.65) & -4 [-10;10] & 52.94 & 47.06 \\ \hline
        education & 0.69 & -2.75 (5.23) & -5 [-8;4] & 62.5 & 37.5 \\ \hline
        financial & 0.61 & -0.29 (7.61) & -4 [-10;10] & 57.14 & 42.86 \\ \hline
        lifestyle & 0.43 & 9.6 (0.89) & 10 [8;10] & 0.00 & 100.00 \\ \hline
        friendship & 0.26 & 1.33 (9.87) & 6 [-10;8] & 33.33 & 66.67 \\ \hline
    \end{tabular} \caption{\label{tab:example}Summary of the life events data collected during the third wave with intensity (M, SD, Median) and proportion of participants who indicated a positive and negative intensity, between -10 (very negative) to +10 (very positive)}
    \label{t5}
\end{table}

\subsubsection*{Stressors during crisis}
To measure psychological stressors, we used a part of the Crisis Coping Assessment Questionnaire (CCAQ) \cite{ccaq}. Specifically, we asked several items from two perspectives: how \textit{worried} they were about a range of concerns over the past two years and how \textit{problematic} each of the concerns turned out to be. For each perspective, participants answered on a 9-point scale (1=did not worry me at all/not problematic at all; 9=worried me extremely/extremely problematic) to the following 12 concerns: their own physical health, mental health, and safety, the physical and mental health and safety of people they love, losing their job, not having enough money to survive, getting basic everyday things (food, etc.), social unrest, separation from their family, a close person being violent.

Responses to the CCAQ showed that participants were most worried about the physical safety and mental health of their loved ones. The extent to which these stressors occurred in reality showed that participants' own mental health and that of their loved ones were impacted (Table \ref{t6}). For all concerns measured, the worry score was never exceeded by the actual problem score. That is, participants were consistently more worried about an issue than that it turned out to be a problem. We see that the worry-problem discrepancy is not evenly distributed across concerns and below, we provide evidence for two sub-groups of participants.

\begin{table}[!ht]
    \centering
    \begin{tabular}{|l|l|l|l|l|}
    \hline
        \textbf{Variable} & \textbf{$M_{worry}$} & \textbf{$SD_{worry}$} & \textbf{$M_{actual}$} & \textbf{$SD_{actual}$} \\ \hline
        Own physical safety & 5.16 & 2.23 & 3.58 & 2.34 \\ \hline
        Own mental state & 5.69 & 2.40 & 4.92 & 2.63 \\ \hline
        Own safety & 4.73 & 2.24 & 2.72 & 1.96 \\ \hline
        Physical safety loved ones & 6.85 & 1.94 & 4.33 & 2.45 \\ \hline
        Mental health loved ones & 6.31 & 2.09 & 4.77 & 2.40 \\ \hline
        Safety loved ones & 6.22 & 2.20 & 3.57 & 2.36 \\ \hline
        Losing job & 3.45 & 2.52 & 2.45 & 2.24 \\ \hline
        Financial problems & 5.06 & 2.59 & 3.88 & 2.63 \\ \hline
        Getting basics & 4.65 & 2.32 & 3.55 & 2.33 \\ \hline
        Social unrest & 4.45 & 2.14 & 3.10 & 2.13 \\ \hline
        Being separated from family & 5.13 & 2.63 & 4.13 & 2.65 \\ \hline
        Violence close person & 1.67 & 1.49 & 1.44 & 1.30 \\ \hline
    \end{tabular}
    \caption{\label{tab:example}Summary of worries about psychological stressors and how problematic each stressor turned out to be (M, SD) on a scale of 1-9 (1=did not worry me at all/not problematic at all; 9=worried me extremely/extremely problematic)}
    \label{t6}
\end{table}

\section*{Data Records}
The RW3D dataset is available on the Open Science Framework at \hyperlink{https://doi.org/10.17605/OSF.IO/6S2JP}{https://doi.org/10.17605/OSF.IO/6S2JP}. The corresponding repository (\hyperlink{https://osf.io/9b85r/}{https://osf.io/9b85r/}) contains all supplementary materials and a variable code book with detail and naming conventions for the full dataset.


\section*{Technical Validation}

This section describes i) the steps taken to ensure data quality through participant exclusion criteria and ii) how data derivatives were obtained.

\subsection*{Data retention and exclusion}
After each wave of data collection, we excluded participants based on two text-based criteria: if the long text was not written in the English language (as determined with the \textit{cld} R package \cite{cld3}) or contained more than 20\% punctuation tokens, participants were excluded. Both criteria were deemed necessary to ensure text data quality. The retention was 70.30\% and 67.13\% in the second and third wave, respectively. Table\ref{t1} shows the sample size and key sample descriptives over the three waves.

\subsection*{Data derivatives}

We obtained two kinds of derivatives from the data, one based on the text data and the other on the emotion and CCAQ questionnaires. From the text data, we arrived at higher-order topics that provide an overarching theme for each written text and can be used to study what participants are writing about. The psychological variables (emotion scales and CCAQ) were mapped to higher-order psychological constructs characterised by clusters of participants on the emotion transition (from 2020 to 2021 and from 2021 to 2022) as well as the worry-problem discrepancy.

\subsubsection*{Topics}
To capture overarching themes in the text data, we constructed a correlated topic model using the \textit{stm} R package \cite{stm} for the text data for each data collection wave. This probabilistic model is based on the assumption that a piece of text consists of a mix of topics, which in turn are a mix of words with probabilities of belonging to a topic \cite{blei2006dynamic, stm}. Table \ref{t4} shows the top three most prevalent topics per wave for the long texts (see supplementary materials Table 2 and 3 for a full list of topics and terms for long and short texts). We have assigned labels to each topic based on the most frequent terms per topic.

\begin{table}[!ht]
    \centering
    \begin{tabular}{|l|l|l|l|}
    \hline
        \textbf{wave} & \textbf{topic} & \textbf{\% documents} & \textbf{terms} \\ \hline
        wave 1 & rule following & 10.29 & peopl, feel, see, mani, die, govern, rule, think, will, follow \\ \hline
        wave 1 & how long will this last? & 10.16 & will, worri, feel, famili, normal, back, long, know, life, hope \\ \hline
        wave 1 & worry about loved ones & 9.53 & worri, also, famili, friend, time, anxious, work, home, feel, will \\ \hline
        wave 2 & hope for normality & 15.04 & will, feel, normal, hope, back, get, vaccin, thing, look, forward \\ \hline
        wave 2 & missing normality & 14.38 & work, want, friend, famili, feel, see, miss, time, abl, home \\ \hline
        wave 2 & anxiety & 11.91 & will, worri, vaccin, concern, also, feel, covid, anxious, effect, virus \\ \hline
        wave 3 & normal life & 9.99 & normal, back, now, life, live, covid, worri, feel, get, can \\ \hline
        wave 3 & still worried & 8.70 & covid, feel, still, worri, get, peopl, test, though, affect, don’t \\ \hline
        wave 3 & new variants & 8.67 & still, variant, vaccin, will, case, virus, new, concern, number, peopl \\ \hline
    \end{tabular}
     \caption{\label{tab:example}Top 3 most prevalent topics for the long texts in each wave, with assigned labels and most frequent terms.}
     \label{t4}
\end{table}

\subsubsection*{Higher-order psychological clusters}
\begin{figure}[!ht]
\centering
\includegraphics[width=\linewidth]{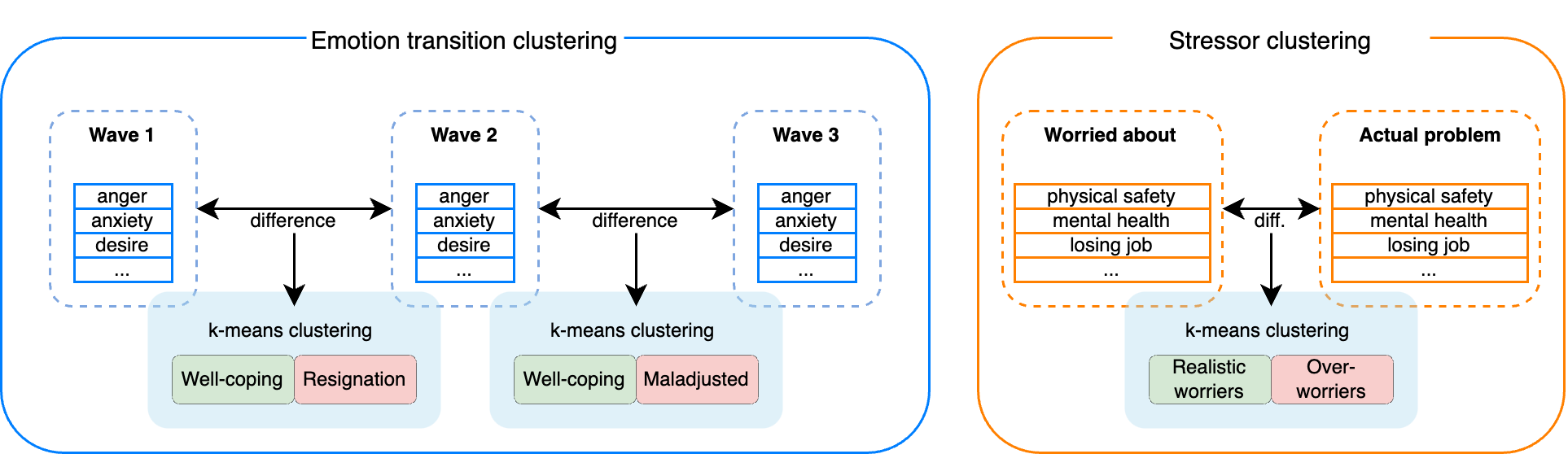}
\caption{Clustering approach for two emotion transitions and for the worry-problem discrepancy. The k-means algorithm showed an ideal number of clusters for $k=2$ as per the Elbow method and the Silhoutte method.}
\label{fig3}
\end{figure}

Earlier work found evidence for hidden clusters within the data in the emotion transition from wave 1 to wave 2 \cite{mozes2021repeated}. We assessed whether there were additional emotion clusters in this extended dataset and also in the discrepancy between the worry and the actual problem posed by stressors in the pandemic. For each concept, we proceeded as follows (Fig. \ref{fig3}): we took the delta value of the emotion ratings for two transitions (wave 2 minus wave 1, and wave 3 minus wave 2) and used the delta between the CCAQ worry rating and the problem rating (worry minus actual problem). For emotion transition 1 (2020 to 2021), transition 2 (2021 to 2022) and the worry-problem discrepancy, we then ran k-means clustering \cite{likas2003global}. We decided on the number of clusters through convergence of the scree plot and the Silhouette method. For all three transition periods, there was evidence of two clusters.

The emotion clusters (Table \ref{t7}) for the 2020-2021 transition were characterised by one group of participants (44.36\% of the sample) that showed a marked improvement in emotional well-being, while another group (55.64\%) showed emotional responses resembling resignation (i.e., these participants reported higher anger, disgust and sadness but also lower worry and fear). The subsequent year’s cluster showed again a well-coping group of participants (40.45\%) with a very similar pattern to the earlier well-coping cluster, with the exception of no increase in desire. This was juxtaposed with a maladjusted group of participants (59.55\%) who overlapped somewhat with the resignation cluster but which we termed differently due to this group’s increase in fear and anxiety and decrease in desire.

\begin{table}[!ht]
    \centering
    \begin{tabular}{|l|l|l|l|l|}
    \hline
        Emotion & 2021 well-coping & 2021 resignation & 2022 well-coping & 2022 maladjusted \\ \hline
        anger & -1.68 & 0.83 & -2.38 & 0.56 \\ \hline
        anxiety & -3.16 & -0.11$^{ns}$ & -2.82 & 0.30 \\ \hline
        desire & 1.24 & 0.25 & 0.27$^{ns}$ & -0.52 \\ \hline
        disgust & -1.26 & 0.53 & -1.74 & 0.51 \\ \hline
        fear & -3.53 & -0.53 & -2.52 & 0.17 \\ \hline
        happiness & 1.85 & -0.14$^{ns}$ & 1.65 & -0.34 \\ \hline
        relaxation & 2.21 & -0.48 & 2.24 & -0.53 \\ \hline
        sadness & -2.77 & 0.56 & -3.27 & 0.28 \\ \hline
        worry & -2.40 & -0.97 & -1.79 & -0.62 \\ \hline
        Size & 44.36\% & 55.64\% & 40.45\% & 59.55\% \\ \hline
    \end{tabular}
   \caption{\label{tab:example} Means per emotion for the emotion transition clusters at the 2020-2021 and 2021-2022 transition (all sign. different from 0 at $p<.01$, two-sided, except for those with \textit{ns}). The two wellcoping clusters show similar patterns that suggest an increase in positive and a decrease in negative emotions. The resignation (2021) and mal-adjusted cluster show some overlap but differ in their change of desire and fear.}
   \label{t7}
\end{table}
With regards to the worry-problem discrepancy clustering (Table \ref{t8}), the larger of the two clusters (58.16\%) was characterised by a markedly stronger “over-worry” (i.e., they indicated to worry about the various concerns much more than that they turned out to be a problem). These over-worriers were particularly evident on questions about the physical health and safety of loved ones. In contrast, the group that we termed the realistic worriers (41.84\%) show consistently lower worry-problem discrepancies. Merely on the questions about someone close being violent (domestic violence) both groups were in agreement (low worry and low actual concern).

\begin{table}[!ht]
    \centering
    \begin{tabular}{|l|l|l|}
    \hline
        Variable & Realistic worriers & Over-worriers \\ \hline
        Own physical safety & 0.76 & 2.74 \\ \hline
        Own mental health & 0.35 & 1.37 \\ \hline
        Own safety  & 1.17 & 3.17 \\ \hline
        Physical health loved ones & 1.21 & 4.33 \\ \hline
        Mental health loved ones & 0.69 & 2.72 \\ \hline
        Safety loved ones & 1.35 & 4.47 \\ \hline
        Losing job & 0.59 & 1.57 \\ \hline
        Financial problems & 0.66 & 1.91 \\ \hline
        Getting basic needs & 0.53 & 1.88 \\ \hline
        Social unrest & 0.85 & 2.04 \\ \hline
        Separation from family & 0.54 & 1.64 \\ \hline
        Domestic violence & 0.22 & 0.24 \\ \hline
        Size & 41.84\% & 58.16\% \\ \hline
    \end{tabular}
    \caption{\label{tab:example}Clustering on the worry-problem discrepancy (all sign. different from 0 at $p<.01$, two-sided)}
    \label{t8}
\end{table}

\section*{Usage Notes}
Understanding and addressing the psychological impact of the COVID-19 pandemic, and possibly preparing for the impact of future global crises, remains an ongoing research challenge. One of the impediments is high quality data that connects different modalities of how individuals experienced the pandemic. The current dataset paper introduced the RW3D, a repeated-measures dataset of UK participants, combining psychological variables examined via survey methods with rich textual responses. The explanatory relationships of coping in the pandemic are yet poorly understood. With the RW3D, we can examine via panel models to what extent life events, concerns raised in text data or socio-demographics changes (e.g., job loss) and variables (e.g., gender), explain transitions into the various coping styles. Gaining insights into these complex relationships could also be a way forward to target interventions at those who most need it. Importantly, the inclusion of various control variables in the dataset allows researchers to control for potential confounds.

Moreover, we can also learn about some fundamental aspects of the relationship between text data and psychological variables. By connecting the modalities, we can test to what extent ground truth emotions are predictable from text data and whether a lagged design can help anticipate emotion changes at a later moment based on text data in previous years. Similarly, since we know about participants’ life events and stressors, we can assess how these are retrievable from the text data. One implicit assumption of plenty of applied text-based research is that these psychological variables are apparent from text data, but rich datasets that allow for such a mapping are scarce.

\subsection*{Limitations}
Some limitations need to be considered when making use of this dataset. First of all, the data were collected from UK participants only. While this allows for a rich analysis of the UK due to country-specific circumstances (e.g., infection spread, government responses), the results may not be generalisable to other populations. Second, some variables were collected retrospectively requiring participants to report significant life events up to 2 years after they happened and to think back about worries and actual problems of crisis coping. Third, while there is considerable spread in demographics, the dataset does not make use of nationally representative sample. A way to mitigate that concern post-hoc might be to weigh sample characteristics according to their prevalence in the UK population (see \cite{bradley2021unrepresentative}).

\section*{Code availability}
There is no custom code associated with this data descriptor. For data(pre)processing and obtaining data derivatives, we used existing R packages. This included \textit{cld} for English language checks \cite{cld3}, \textit{quanteda} \cite{quanteda} and \textit{stringr} \cite{stringr} for text metadata (number of characters, tokens, punctuation), the \textit{stm} package \cite{stm} for constructing topic models, and the \textit{factoextra} package \cite{factoextra} for the determination of the number of clusters for obtaining the higher-order psychological constructs.

\bibliography{sample,datasets}

\end{document}